# The use of invariant moments in hand-written character recognition

**Dan L. Lacrămă**
"Tibiscus" University of Timişoara, Romania
**Ioan Şnep**
NMM Paris Branch, France

**ABSTRACT:** The goal of this paper is to present the implementation of a Radial Basis Function neural network with built-in knowledge to recognize hand-written characters. The neural network includes in its architecture gates controlled by an attraction/repulsion system of coefficients. These coefficients are derived from a preprocessing stage which groups the characters according to their ascendent, central, or descendent components.
The neural network is trained using data from invariant moment functions. Results are compared with those obtained using a K nearest neighbor method on the same moment data.

## 1. Introduction

### 1.1. Optical Hand-written Character Recognition

Despite the widespread use of computers, an ever increasing amount of data is obtained in hand-written form on paper. Consequently, there is a great demand for an automatic system capable of transforming this data into machine readable form, accurately and quickly [Bre97].

In order to recognize and transcribe the hand-written input contained in a complete field image, two problems need to be solved:

- the image of the document must be segmented into rows, words and finally into individual characters
- the individual characters need to be recognized

The document segmentation into rows and words is currently solved quite satisfactorily, but this is not the case with the segmentation of the word into characters [S+93]. A lot of work is currently done worldwide to solve this difficult task.

In our research the word segmentation into characters will not be considered - the research concentrates on the recognition of characters. A technique based on vertical division is used for recognition of the individual characters [HHS92, LA96] as shown in Figure 1.1.

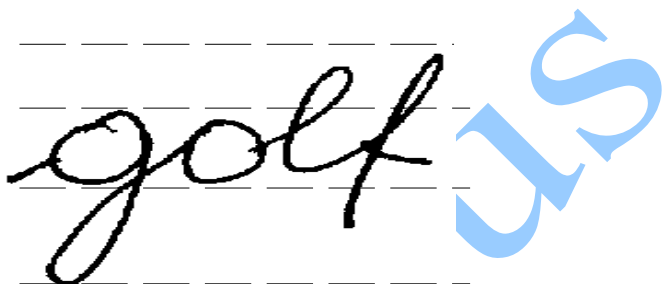

*Fig. 1.1 Vertical classification of characters*

This approach leads to a clustering of the alphabet's lower case symbols into four *categories*:

- Central letters (e.g. a, c, e ...)
- Ascendant letters (e.g. b, i, t ...)
- Descendent letters (e.g. g, p, y ...)
- Ascendant & descendent (e.g. f, j ...)

The analysis also uses features like strokes, breaks or closed and opened loops to further divide these categories into smaller *groups* each containing just a few letters. For example using these features the Central letters category is subdivided in four groups: {a, o}; {c, e, r, s};{m, n, u, v, w} and {x, z}.

Therefore the classification task is itself subdivided into three steps:

- category classification (ascendent etc.)
- group classification (strokes/loops etc.)
- final decision inside group

This paper concentrates on the use of the invariant moments and tree Radial Basis Functions neural network to achieve the final decision between similar characters in the final groups.

## 1.2. Invariant Moments

A set of invariant moments was defined by Hu as f(x,y) image descriptors invariant to the basic geometrical transformations: translation rotation and scale [Hu62].

$$
\begin{aligned}
H_1 &= n_{20} - n_{02} \\
H_2 &= (n_{20} - n_{02})^2 + 4 \cdot n_{11}^{2} \\
H_3 &= (n_{30} - 3 \cdot n_{12})^2 + (3 \cdot n_{21} - n_{03})^2 \\
H_4 &= (n_{30} + n_{12})^2 + (n_{21} + n_{03})^2
\end{aligned}
\tag{1.2.1}
$$

where $n_{pq}$ is the (p+q) order normalized central moment with:

$$
m_{pq} = \iint\limits_{A} x^p y^q f(x,y) \cdot dx \cdot dy
\tag{1.2.2}
$$

$$
n_{pq} = \frac{m_{pq}}{m^{\gamma}{}_{00}} \text{ and } \gamma = 1 + \frac{p+q}{2}
\tag{1.2.3}
$$

The *Hu moments* are quite sensitive to noise and their invariance is only available in a continuous image. In the case of hand-written character recognition it is vital to improve the invariance to digitization and the noise effect of individual handwriting .

The *Zernike moments* are less sensitive to noise [KH90]. The first six of them can be computed from $m_{pq}$ moments:

$$Z_{00} = \frac{m_{00}}{\pi}$$

$$Z_{22} = 3 \cdot \frac{m_{02} - m_{20} - 2 \cdot i \cdot m_{11}}{\pi}$$

$$Z_{20} = 3 \cdot \frac{2 \cdot m_{20} + 2 \cdot m_{02} - 1}{\pi}$$

$$Z_{33} = 4 \cdot \frac{m_{03} - 3 \cdot m_{21} + i \cdot (m_{30} - 3 \cdot m_{12})}{\pi}$$

$$Z_{31} = 12 \cdot \frac{m_{03} + m_{21} - i \cdot (m_{30} + m_{12})}{\pi}$$

$$Z_{44} = 5 \cdot \frac{m_{40} - 6 \cdot m_{22} + m_{04} + 4 \cdot i \cdot (m_{31} m_{13})}{\pi} \qquad (1.2.4.)$$

The *Affine moments* are derived to be invariant to affine transformations; rotation translation and angle in an digitized image [FS94]. They also are computed using $m_{pq}$ moments:

$$A_1 = \frac{(m_{20} \cdot m_{02} - m_{11}^2)}{m_{00}^4}$$

$$A_2 = (m_{30}^2 \cdot m_{03}^2 - 6 \cdot m_{30} \cdot m_{21} \cdot m_{12} \cdot m_{03} +$$
$$+ 4 \cdot m_{30} \cdot m_{12}^3 - 4 \cdot m_{03} \cdot m_{21}^3 - 3 \cdot m_{21}^2 \cdot m_{12}^2) / m_{00}^{10}$$

$$A_3 = (m_{20} \cdot (m_{21} \cdot m_{03} - m_{12}^2) -$$
$$- m_{11} \cdot (m_{30} \cdot m_{03} - m_{21} \cdot m_{12}) +$$
$$+ m_{02} \cdot (m_{30} \cdot m_{12} - m_{21}^2)) / m_{00}^7 \qquad (1.2.5.)$$

When used as descriptors of a hand-written character in a binary image, all these three sets are useful but none of them can cover the whole complexity of the classification task.

*1.3. Radial basis function neural networks*

Radial Basis Function (RBF) Neural Networks have been employed quite extensively in the last few years to perform difficult pattern recognition tasks [Hay94].

The optical hand-written character recognition (OHCR) problem and other similar classification tasks are basically solved by transforming them into a high-dimensional space in a nonlinear manner. The underlying justification for so doing is *Cover's theorem on the separability of patterns* [Cov65].

This nonlinear transformation is performed by a RBF neural network:

- The input layer contains a number of source nodes equal to the dimension of the input vector X.
- The hidden layer usually consists of a large number of RBF neurons directly connected to all the input nodes. The transformation from the input to the hidden layer is nonlinear
- The output layer consists of linear neurons connected to all the hidden neurons. The transformation from the hidden layer to the output layer is linear.

The learning strategy applied to the RBF network is dependent on its type [Rip96] (regularization or generalized) :

*Regularization RBF networks* have fixed centers corresponding to the clustering centers of the training set data (sometimes the centers are selected at random from the training set data points). The weights vector W is computed using the matrix equation:

$$W = G^+ \times D \qquad (1.3.1.)$$

where D is the desired response vector and $G^+$ is the pseudoinverse of the matrix G=[$g_{ji}$], $g_{ji}$ being the Green's function

$$g_{ji} = \exp(-\frac{\overline{X}}{\sigma^2} \left\| X_j - T_i \right\|^2)$$

$$j = 1, 2, \ldots, N; \quad i = 1, 2, \ldots, M \qquad (1.3.2.)$$

where $X_j$ is the j-th input of the training set, $T_i$ is the fixed centre position, $\overline{X}$ is the average and $\sigma$ the dispersion

The *Generalized RBF network's* centers and weights undergo a supervised change during the learning process.

- Linear weights (output layer) $w_i$

$$w_i(n+1) = w_i(n) - \eta_1 \frac{\partial E(n)}{\partial w_i(n)} \qquad (1.3.3.)$$

- Position of centres (hidden layer)

$$T_i(n+1) = T_i(n) - \eta_2 \frac{\partial E(n)}{\partial T_i(n)} \qquad (1.3.4.)$$

with E(n)=a computed error function at time n.

Reported use of RBF Neural Networks in hand-written character recognition show encouraging results, but the large number of RBF neurons in the hidden layer usually restricts the applicability of the method to the recognition of hand-written numerals. [Lee99, Lem93].

## 2. Implementation of the tree RBF neural network for optical character recognition

Building an RBF neural network capable of dealing with all the alphabet letters requires a huge hidden layer, and consequently large matrices calculations.

Developing the research on the three stages described above, it became clear that the most significant part (more that 80% in each of the testing sets) of the classification errors occurred in the last stage when the delicate decision among very similar letters must be taken.

Taking also account of the fact that at the end of each stage the field of possible classification choices tightens, the idea of a tree structure neural network originated.

During the implementation of this network the possibility arises of blocking the inactive branches.

These partial obstructions are done by a set of validation gates each controlled by an attraction/repulsion signal $C_{A/R}$ dependent on the decisions in the previous stages.

The implemented structure of the tree network is presented as a block diagram in Figure 2.1.

If the $C_{A/R}$ is '+1' the branch is enabled and the final decision will arise from its output (attraction to the correspondent group). If it is '-1' the branch is disabled and the final decision will not have any components from the corresponding group (repulsion).

These validation control parameters are the outputs of the *Root Decision* where the first two steps of classification are made.

The RBF neural network branch architecture is represented in Figure. 2.2.

Because of the separation made by the use of $C_{A/R}$ the internal parameters of each branch are computed separately. This simplifies the calculations and therefore is less time consuming.

## 3. Experimental Results

As stated in the introductory section each of the three sets of moments can be employed as image descriptors but the classification performance in the case of hand-written characters is not satisfactory.

In the *first stage of the experiments* the K nearest neighbor method was used for ten samples of each letter of the alphabet written by the same subject. The classification results in the case of each moment set are summarized in Table 3.1.

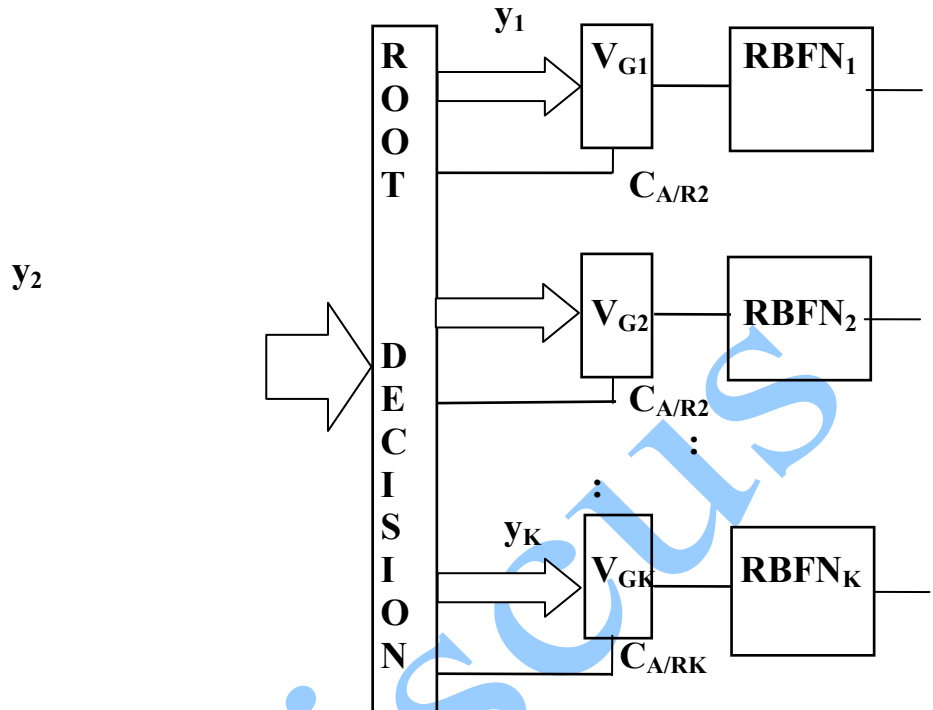

*Fig. 2.1: Tree RBF neural network*

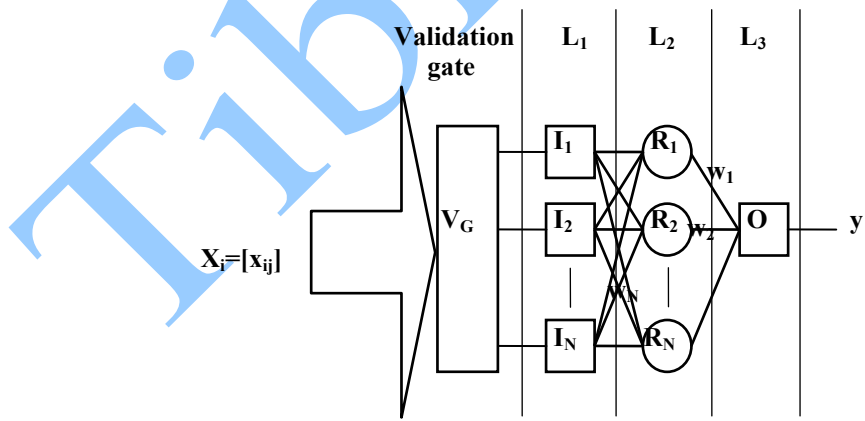

*Fig. 2.2: RBF neural network branch*

***Table 3.1***

| Moments set | Correct Classification [%] |
|-------------|----------------------------|
| Hu | 53.08 |
| Affine | 69.23 |
| Zernike | 72.31 |

The same samples were classified with the same method, but using a mixed description vector containing selected moments from $H_1$-$H_4$, $A_1$-$A_4$ and $Z_1$-$Z_4$. A classification accuracy of 93.75% was obtained.

The selection of the moments was used in order to reduce the amount of computing. It contained: $A_1$, $A_2$, $H_1$, $Z_{00}$ and $Z_{31}$ which proved during the experiments to be the most relevant. (the biggest variations among different clusters and best constancy inside one cluster).

In the *second stage of the experiments* when the RBF network was employed all the moments were used.

For evaluating the behaviour and performance of the RBF tree neural network three kinds of training sets were used:

A. Hand-written characters of a single subject. Ten samples of each of the 26 English alphabet letters were considered. A second group of samples with the same magnitude was used as test set.

B. Hand-written characters of ten different subjects. Five samples of each English alphabet letter were considered (i.e. 1300 samples). The test set had the same composition and contained 500 samples from the C set.

C. Hand written characters of thirty subjects (different from the ones in B set). 1500 samples were selected at random with the only restriction that each alphabet letter must be represented. The test set contained 500 samples also selected at random in the same conditions as the training set.

In each case the selection of the RBF units centers was made based on the clustering centers revealed by a pre-classification inside the training set with the simple K-nearest neighbor method employed as described in the first stage of the experiments.

The tests have shown that all branch parameters can be computed separately without affecting the overall performance of the system.

It is also a positive fact that even in the case C the largest branch contained only 37 RBF units in its hidden layer.

Errors in classification of similar characters (i.e. characters of the same group) were reduced with 89.21% in set A, 47.42% in case B and 23.84% in case C. Overall results are given for the three sets in Table 3.2

*Table 3.2.*

| Characters set | Correct Classification [%] |
|----------------|----------------------------|
| A | 95.36 |
| B | 88.60 |
| C | 78.40 |

The use of different additional input data for each branch is in course of being implemented.

## Conclusions

The most obvious concluding remark of the above result is that it is a significant improvement from the starting stage and this was achieved without employing large structures and extensive time-consuming computations.

Improvements in performance should come from using a more flexible structure to reduce the effect of root decision errors (i.e. fuzzification of the $C_{A/R}$ parameters).

This can prove useful for another objective to be met: the availability of multi-decisions (i.e. the character is an 'a' with a 78% probability or an 'o' with a 22% probability) as recommended in reference [Bai93].

It is also possible to implement such tree RBF neural networks for other complex tasks in pattern recognition.

## Acknowledgments

The authors wish to thank The University of Central Lancashire Preston for the technical and documentary resources made available for the research on hand-written recognition.